\documentclass{article}

\usepackage{microtype}
\usepackage{graphicx}
\usepackage{subcaption}
\usepackage{booktabs} 

\usepackage{hyperref}

\usepackage[preprint]{icml2026}

\usepackage{amsmath}
\usepackage{amssymb}
\usepackage{mathtools}
\usepackage{amsthm}

\usepackage[capitalize,noabbrev]{cleveref}

\theoremstyle{plain}

\theoremstyle{definition}

\theoremstyle{remark}

\usepackage[textsize=tiny]{todonotes}

\icmltitlerunning{Beyond Scaling: Assessing Strategic Reasoning and Rapid Decision-Making Capability of LLMs in Zero-sum Environments}

\begin{document}

\twocolumn[
  \icmltitle{Beyond Scaling: Assessing Strategic Reasoning and Rapid Decision-Making Capability of LLMs in Zero-sum Environments}



  \icmlsetsymbol{equal}{*}

  \begin{icmlauthorlist}
    \icmlauthor{Yang Li}{tsinghua}
    \icmlauthor{Xing Chen}{} 
    \icmlauthor{Yutao Liu}{ouc}
    \icmlauthor{Gege Qi}{caict}
    \icmlauthor{Yanxian BI}{caeit}
    \icmlauthor{Zizhe Wang}{buaa}
    \icmlauthor{Yunjian Zhang}{ucas}
    \icmlauthor{Yao Zhu}{zju}
  \end{icmlauthorlist}

  \icmlaffiliation{tsinghua}{Tsinghua University}
  \icmlaffiliation{ouc}{Ocean University of China (OUC)}
  \icmlaffiliation{caict}{CAICT}
  \icmlaffiliation{caeit}{CAEIT}
  \icmlaffiliation{buaa}{Beihang University}
  \icmlaffiliation{ucas}{UCAS}
  \icmlaffiliation{zju}{Zhejiang University}

  \icmlcorrespondingauthor{}{}

  \icmlkeywords{Machine Learning, ICML}

  \vskip 0.3in
]



\printAffiliationsAndNotice{}  

\begin{abstract}
Large Language Models (LLMs) have achieved strong performance on static reasoning benchmarks, yet their effectiveness as interactive agents operating in adversarial, time-sensitive environments remains poorly understood. Existing evaluations largely treat reasoning as a single-shot capability, overlooking the challenges of opponent-aware decision-making, temporal constraints, and execution under pressure. This paper introduces Strategic Tactical Agent Reasoning (STAR) Benchmark, a multi-agent evaluation framework that assesses LLMs through 1v1 zero-sum competitive interactions, framing reasoning as an iterative, adaptive decision-making process. STAR supports both turn-based and real-time settings, enabling controlled analysis of long-horizon strategic planning and fast-paced tactical execution within a unified environment. Built on a modular architecture with a standardized API and fully implemented execution engine, STAR facilitates reproducible evaluation and flexible task customization. To move beyond binary win–loss outcomes, we introduce a Strategic Evaluation Suite that assesses not only competitive success but also the quality of strategic behavior, such as execution efficiency and outcome stability. Extensive pairwise evaluations reveal a pronounced strategy–execution gap: while reasoning-intensive models dominate turn-based settings, their inference latency often leads to inferior performance in real-time scenarios, where faster instruction-tuned models prevail. These results show that strategic intelligence in interactive environments depends not only on reasoning depth, but also on the ability to translate plans into timely actions, positioning STAR as a principled benchmark for studying this trade-off in competitive, dynamic settings. Codes are available at \url{https://github.com/star-nexus/star}.
\end{abstract}

\begin{figure*}[t]
  \centering
  \includegraphics[width=0.85\linewidth]{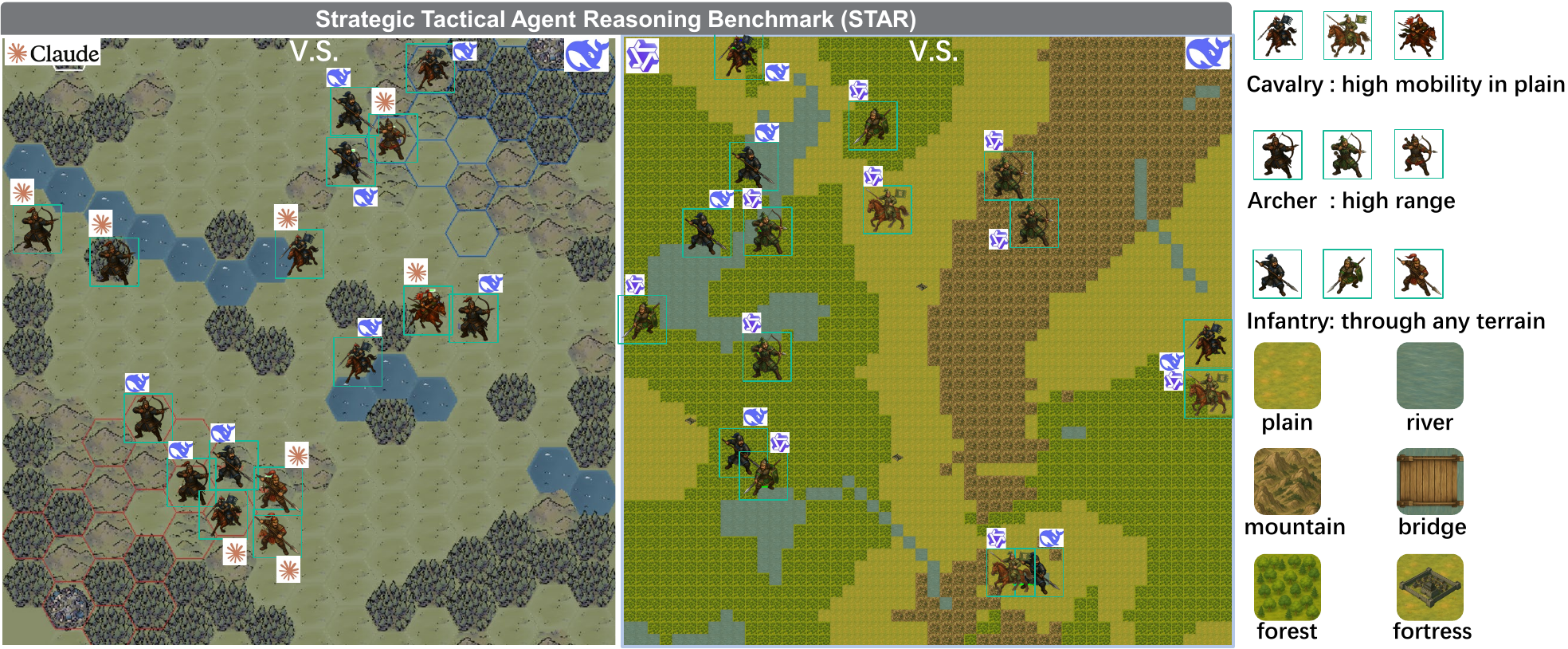}
  \caption {Visualization interface of STAR (Strategic Tactical Agent Reasoning Benchmark). STAR is a LLM PvP(Player versus Player) environment with multiple types of maps, evaluating the capability of LLMs to perform iterative reasoning and strategic decision-making in dynamic and zero-sum multi-agent environments.}
    \label{fig:presentation}
\end{figure*}

\section{Introduction}
Large Language Models (LLMs) have achieved remarkable progress in recent years \cite{dubey2024llama,qwen2,zhu2025patchwise}, driven by large-scale pretraining and advances in reasoning-oriented techniques such as Chain-of-Thought prompting \cite{wei2022chain} and architectural optimization \cite{lu2024deepseekvl}. As a result, modern LLMs demonstrate strong performance across a wide range of static benchmarks, including mathematical reasoning \cite{cobbe2021training}, code generation \cite{ni2024l2ceval}, and factual knowledge evaluation \cite{wang2024mmlu}. These successes have established reasoning as one of the central capabilities in contemporary language models.

However, most existing evaluations implicitly treat reasoning as a single-shot, context-independent process: the model observes a fixed input, produces an answer, and is scored in isolation. Such settings fail to reflect scenarios where reasoning must unfold over time, react to an opponent, and adapt to continuously evolving states. In particular, they overlook a critical dimension of intelligence: strategic reasoning under adversarial interaction, where outcomes depend not only on the correctness of individual decisions, but also on anticipation, adaptation, and long-horizon planning. As LLMs are increasingly used as decision-making agents, this gap between static evaluation and interactive reasoning becomes increasingly pronounced. A closer examination reveals several limitations shared by existing benchmarks. First, most evaluations focus on single-agent performance in non-adversarial environments, providing little insight into a model’s ability to reason about and respond to competing agents. Second, temporal structure is largely ignored: models are typically allowed unconstrained inference time, masking the trade-off between reasoning depth and decision timeliness that arises in sequential decision-making. Third, evaluation metrics are often coarse-grained (e.g., accuracy or success rate), making it difficult to disentangle how a model wins or loses, and whether its reasoning leads to robust strategic outcomes.

To address these limitations, we introduce the Strategic Tactical Agent Reasoning Benchmark (STAR), a multi-agent evaluation framework designed to assess LLMs through 1v1 competitive interactions in a zero-sum environment. STAR formulates strategic reasoning as an iterative decision-making process, where two agents repeatedly interact, observe each other’s actions, and adapt their strategies over time. This design enables direct measurement of a model’s ability to perform opponent-aware reasoning, maintain situational consistency, and plan over extended horizons, where capabilities that are difficult to capture in static benchmarks.

Architecturally, STAR is organized into four decoupled layers, including framework, environment, communication, and agent, which together provide a clean separation between game logic, interaction protocols, and model-specific behaviors. Importantly, STAR includes a fully implemented underlying execution environment, allowing users to define new tasks and rules on top of the existing infrastructure without reimplementing core mechanics. Through a unified interaction interface and open API, LLMs can be seamlessly deployed as autonomous agents across different task instantiations. Beyond empirical win rates, STAR incorporates a strategic evaluation suite named Performance-Weighted ELO Rating (PWER) tailored for multi-turn adversarial settings, capturing not only competitive outcomes but also the efficiency and stability of strategic execution. By grounding evaluation in interactive competition and supporting user-defined tasks, STAR functions not only as a benchmark but as a general-purpose platform for studying strategic reasoning and agent behavior in dynamic, adversarial environments.

We conduct extensive pairwise evaluations of a diverse set of representative LLMs on the STAR benchmark through systematically designed 1v1 competitive experiments. The results reveal a clear correlation between a model’s reasoning quality and its competitive performance, while simultaneously exposing critical failure modes that are invisible in static evaluations. Notably, models that excel in isolated reasoning tasks do not necessarily translate this capability into superior outcomes in adversarial interactions, underscoring that strategic reasoning must be understood as a dynamic, agent-centric process rather than a static competence. Overall, these findings demonstrate that STAR can effectively differentiate models based on their ability to convert reasoning sophistication into actionable strategic advantage, providing a complementary and necessary perspective for evaluating the strategic reasoning capabilities of modern LLMs. Figure~\ref{fig:presentation} illustrates a representative encounter between two LLM agents, highlighting their decision-making behaviors in a dynamic competitive setting.

Our contributions are summarized as follows:

\begin{itemize}
\item We introduce STAR, the first benchmark that evaluates Large Language Models in explicit zero-sum, 1v1 competitive environments, where reasoning quality is assessed through direct adversarial interaction rather than static, single-agent prompts. By framing reasoning as an iterative, opponent-aware decision process, STAR enables systematic evaluation of strategic consistency, adaptation, and long-horizon planning capabilities that are fundamentally inaccessible to existing static benchmarks.
\item We design a layered and modular agentic framework for strategic evaluation, consisting of decoupled framework, environment, communication, and agent layers. STAR provides a fully implemented execution environment and a unified interaction API, allowing users to easily define new tasks and deploy heterogeneous LLM-based agents, thereby ensuring scalability, extensibility, and fair comparison across models.
\item We propose a multi-dimensional strategic evaluation suite tailored to adversarial settings, which goes beyond empirical win rates to quantify execution efficiency and strategic stability during interaction. This evaluation paradigm exposes critical gaps between abstract reasoning ability and effective action under time and adversarial pressure, offering fine-grained diagnostic insights into how and why LLMs succeed or fail as interactive agents.
\end{itemize}

\begin{figure*}[t]
  \centering
  \includegraphics[width=.825\linewidth]{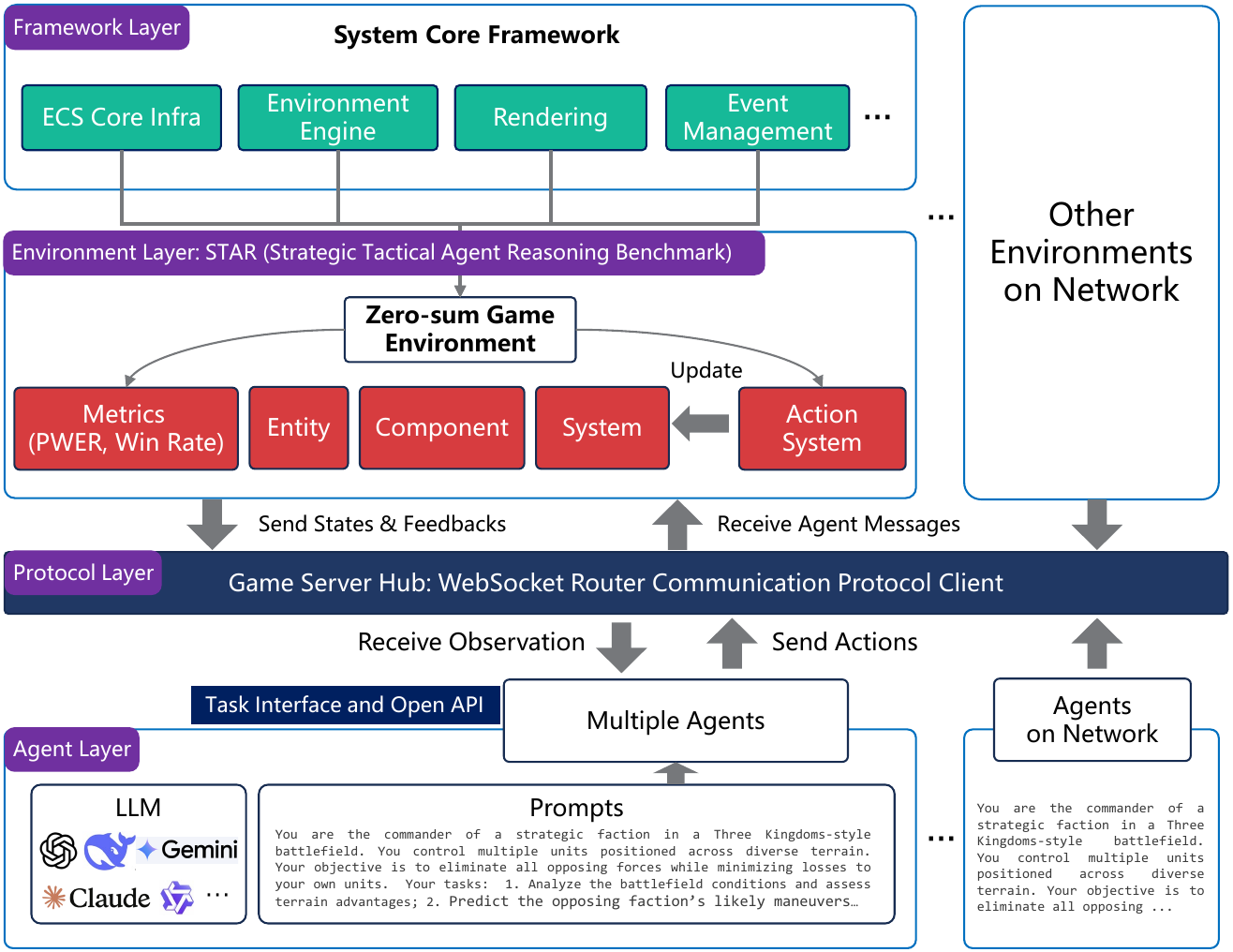}
  \caption {
  Overview of the STAR benchmark architecture. The system relies on four decoupled layers designed for high extensibility and interoperability. The \textbf{Framework Layer} serves as the foundational ECS engine, empowering researchers to construct diverse game scenarios in the \textbf{Environment Layer} by reusing core simulation components. Crucially, the \textbf{Protocol Layer} establishes a standardized interface that bridges the \textbf{Agent Layer} with the environment, enabling seamless interconnectivity between heterogeneous agents and any environment built upon the framework.
  }
    \label{fig:framework}
\end{figure*}

\section{Related Work}

Benchmarks play a vital role in LLM research by revealing model limitations and guiding future advancements. Existing evaluation benchmarks can be roughly categorized based on the specific model capabilities they focus on, such as knowledge and language understanding \cite{wang2018glue,hendrycksmeasuring}, tool learning \cite{huangmetatool,liu2025wtu,wang2024gta}, code generation \cite{chen2021evaluating,li2022competition}, reasoning capability \cite{cobbe2021training,liu2024mathbench}, and so on.

\textbf{Knowledge and language understanding:} The early benchmarks GLUE series \cite{wang2018glue,wang2019superglue} establish a foundation for evaluating the language understanding capabilities of LLMs. Building on this, later benchmarks such as MMLU \cite{hendrycksmeasuring} and BigBench \cite{srivastavabeyond} significantly expand the evaluation scope, covering 57 and 204 diverse tasks, respectively. However, \citet{sanh2022multitask} noted that the standard four-option multiple-choice format widely used in these benchmarks may not sufficiently challenge LLMs or effectively distinguish models with similar performance, leading to potentially inflated results. To address this limitation, MMLU-Pro \cite{wang2024mmlu} increases the number of answer choices from four to ten, improving the benchmark’s ability to differentiate model performance and substantially reducing the likelihood of correct answers being chosen by chance.

\textbf{Tool learning:} MetaTool \cite{huangmetatool} and WTU-Eval \cite{liu2025wtu} focus on assessing LLMs’ ability to recognize when tools are needed and to select appropriate tools based on user needs. In contrast, APIBench \cite{patil2024gorilla}, ToolBench \cite{xu2023tool}, and API-BLEND \cite{basu2024api} place greater emphasis on evaluating the accuracy of tool selection and parameter configuration. These benchmarks often rely on model-generated queries, which may not reliably reflect real user needs. Accordingly, GTA \cite{wang2024gta} enhances the practicality of evaluation by incorporating real-world tools and generating queries grounded in actual user needs.

\textbf{Code generation:} HumanEval \cite{chen2021evaluating}, MBPP \cite{austin2021program}, and L2CEval \cite{ni2024l2ceval} concentrate on natural language-to-Python code generation tasks, assessing the capability of LLMs to produce functional code from given instructions. MBXP \cite{athiwaratkunmulti} broadens this evaluation framework to encompass multiple programming languages. CodeContests \cite{li2022competition} specifically evaluates the performance of LLMs on challenging algorithmic problems typical of competitive programming.

\textbf{Reasoning capability:} GSM8K \cite{cobbe2021training} assesses the multi-step mathematical reasoning via 8,500 diverse grade-school math word problems. MathBench \cite{liu2024mathbench} covers a broad range of mathematical problems, progressing from basic arithmetic to college-level mathematics. NumGLUE \cite{mishra2022numglue} includes both numerical computation and broader reasoning challenges. AGIEval \cite{zhong2024agieval} evaluates foundation models in human-centric standardized exams, including college entrance tests, math competitions, and legal qualification tests. CRASS \cite{frohberg2022crass} focuses on assessing counterfactual reasoning by introducing counterfactual conditional statements as evaluation tools.

Existing benchmarks are limited to assessing the performance of single models on static tasks, overlooking interaction with dynamic environments and counterparts. Consequently, they fail to effectively measure a model’s ability to reason under evolving conditions, adapt to unforeseen changes, and plan over extended decision sequences, which are crucial for intelligent behavior in interactive and competitive scenarios. To bridge this gap, this paper introduces STAR, a benchmark built on zero-sum multi-agent game environments, offering a new lens for evaluating LLM capabilities.

\section{Strategic Tactical Agent Reasoning (STAR) Benchmark}
The STAR benchmark is designed to rigorously evaluate the strategic reasoning and rapid planning capabilities of LLMs in zero-sum, competitive multi-agent game environments. This section first introduces the design principles and construction of zero-sum tasks, followed by a comprehensive overview of the STAR framework. We then describe the underlying inference mechanism of STAR.

\subsection{Task Formalization}

We formalize the strategic reasoning task in STAR as a finite-horizon, partially observable, zero-sum stochastic game. Let the game be defined by the tuple:
\begin{equation}
\mathcal{G} = \langle \mathcal{N}, \mathcal{S}, \{\mathcal{A}_i\}_{i \in \mathcal{N}}, \mathcal{T}, \{\mathcal{O}_i\}_{i \in \mathcal{N}}, \mathcal{R}, H \rangle,
\end{equation}
where $\mathcal{N}$ denotes the set of agents, $\mathcal{S}$ the global environment state, and $\mathcal{A}_i$ the action space of agent $i$. The transition function $\mathcal{T}: \mathcal{S} \times \mathcal{A}_1 \times \dots \times \mathcal{A}_N \rightarrow \Delta(\mathcal{S})$ governs the (potentially stochastic) environment dynamics.
At each timestep, agents receive private observations $o_i \in \mathcal{O}_i$ determined by a vision-based Fog of War mechanism that filters the underlying Entity-Component-System (ECS) state, rendering the game partially observable. Agents select actions based on their observation histories, and the environment evolves according to $\mathcal{T}$. The reward function $\mathcal{R}$ is zero-sum across competing teams, with individual agent rewards aligned within each team. The episode terminates after a fixed horizon $H$ or upon satisfying scenario-specific terminal conditions.
STAR provides a system-level abstraction that operationalizes this formal game definition while remaining agnostic to agent architectures.

\subsection{Zero-sum Competitive Task Design}
To systematically evaluate the strategic reasoning capabilities of LLMs in competitive settings, we design a two-player zero-sum wargame-style task. The environment focuses on adversarial decision-making over discrete combat units, emphasizing spatial reasoning, tactical coordination, and long-horizon planning within a controlled and symmetric setting. The battlefield is instantiated as a grid-based map with heterogeneous terrain types, each imposing different constraints on unit movement and positioning. Two opposing agents, each controlled by an LLM, are initialized with identical resources and unit configurations, ensuring a fair and symmetric competitive setting. Agents interact through sequential decision-making under partial observability, requiring them to anticipate opponent actions and adapt strategies in response to evolving battlefield states. This zero-sum formulation induces a challenging sequential decision process that isolates strategic consistency, spatial planning, and tactical execution as the core competencies under evaluation.

\subsection{STAR Framework}

The STAR framework is a modular, hierarchical system architecture for evaluating LLM reasoning in dynamic, zero-sum multi-agent environments. As shown in Fig.~\ref{fig:framework}, it consists of four decoupled layers, including the Framework Layer, Environment Layer, Protocol Layer, and Agent Layer, designed to ensure scalability and extensibility.

\textbf{Framework Layer (Core Engine).} At the foundation lies a robust game engine built upon the Entity-Component-System (ECS) paradigm. Unlike traditional Object-Oriented Programming (OOP) approaches, the ECS architecture decouples state from behavior. \textit{Entities} act as globally unique identifiers; \textit{Components} serve as pure data containers (e.g., coordinates, unit attributes); and \textit{Systems} encapsulate logic (e.g., movement, rendering) that operates on entities possessing specific component signatures. This design promotes strict data-logic separation and dynamic composability, enabling researchers to inject new behaviors or modify simulation rules without altering the underlying entity definitions.

\textbf{Environment Layer (Simulation Logic).} The specific game rules and world dynamics are implemented in the Environment Layer, instantiated here as the \textit{RoTK} scenario. This layer manages the hexagonal grid map, terrain effects, and combat resolution. Crucially, it implements a \textit{Fog of War} mechanism via a Vision System, ensuring a partially observable environment that enforces uncertainty—a key requirement for testing strategic reasoning under information asymmetry.

\textbf{Protocol (Communication Interface).} To support diverse LLM agents without imposing architectural constraints on the engine, STAR employs a standardized, asynchronous communication protocol. Data exchange is managed via WebSocket using a structured ``Envelope'' format, which encapsulates message types (e.g., Observations, Actions), sender/receiver identity, and timestamps. This isolation facilitates the seamless integration of heterogeneous agents, from API-based commercial models to local open-source weights, and supports scalable multi-agent evaluations.

\textbf{Agent Layer (Decision Host).} The top layer serves as the runtime host for LLM agents, implementing a standardized \textit{Perception-Planning-Action} loop. This layer handles the critical translation process: it converts the environment's structured ECS state into optimized natural language prompt contexts, managing context window constraints while preserving essential strategic information. Conversely, it parses the model's outputs into executable protocol directives, bridging the gap between probabilistic LLM reasoning and deterministic game logic.

\subsection{Extensibility and Customization}
A core design philosophy of STAR is its high extensibility. By decoupling the simulation engine from specific game rules, the architecture allows researchers to develop custom environments—ranging from cooperative resource management to complex diplomatic simulations—using the existing ECS infrastructure without reinventing the wheel. Furthermore, the protocol-agnostic design ensures that upgrading agent capabilities or integrating novel multimodal models requires no modification to the underlying environment, significantly lowering the barrier for future research in dynamic multi-agent systems.

\begin{table*}[h]
\centering
\caption{Turn-Based Mode Leaderboard. Models are ranked by PWER.}
\label{tab:turn-based-results}
\begin{tabular}{lccc}
\toprule
\textbf{Model} & \textbf{PWER} & \textbf{SER} & \textbf{Win Rate} \\
\midrule
Kimi-K2-Thinking & 1206.1 $\pm$ 7.3 & 1149.2 $\pm$ 3.7 & 1.00 \\
GLM-4.7 & 1182.7 $\pm$ 9.5 & 1122.6 $\pm$ 5.2 &  0.857 \\
DeepSeek-Chat & 1166.9 $\pm$ 16.8 & 1112.3 $\pm$ 9.5 &  0.812 \\
GLM-4.6 & 1098.6 $\pm$ 14.2 & 1066.4 $\pm$ 7.1 &  0.714 \\
MiniMax-M2.1 & 1078.8 $\pm$ 16.4 & 1053.9 $\pm$ 7.9 &  0.625 \\
Qwen3-32B & 1006.8 $\pm$ 12.0 & 1012.8 $\pm$ 6.8 &  0.538 \\
Qwen3-30B-A3B-Thinking & 1005.6 $\pm$ 11.4 & 998.5 $\pm$ 6.2  & 0.500 \\
GPT-OSS-20B & 988.2 $\pm$ 12.1 & 988.0 $\pm$ 6.2 &  0.462 \\
Qwen3-14B & 972.7 $\pm$ 14.9 & 979.1 $\pm$ 8.5 &  0.385 \\
Qwen3-8B & 925.1 $\pm$ 9.1 & 952.9 $\pm$ 4.3 & 0.300 \\
Qwen3-30B-A3B-Instruct & 877.3 $\pm$ 11.8 & 913.9 $\pm$ 5.2 & 0.231 \\
Nemotron-Nano-9B-v2 & 865.6 $\pm$ 9.5 & 910.9 $\pm$ 4.1 &  0.182 \\
Kimi-K2-Instruct & 834.9 $\pm$ 13.5 & 880.2 $\pm$ 6.5 & 0.143 \\
Qwen3-8B-NoThinking & 790.7 $\pm$ 8.2 & 859.3 $\pm$ 3.9 & 0.077 \\
\bottomrule
\end{tabular}%
\end{table*}

\section{Experimental Results}
\label{sec:results}
\subsection{Settings and Metrics}
We evaluate multiple representative LLMs across two game modes designed to characterize distinct decision-making regimes. The \textbf{Turn-Based mode} focuses on unconstrained reasoning and strategic optimality, whereas the \textbf{Real-Time mode} evaluates the model’s capacity to maintain reasoning quality while operating under real-time constraints, thereby revealing the inherent trade-off between deliberation and responsiveness.

To obtain a comprehensive and fair assessment, each model competes against all other evaluated models in a round-robin manner, and match outcomes are recorded for every pairing. Given this evaluation protocol, we employ three complementary metrics that capture different aspects of competitive performance: Win Rate, Standard ELO Rating (SER), and Performance-Weighted ELO Rating (PWER). The win rate provides a direct and interpretable measure of pairwise competitiveness, reflecting how often a model prevails against its opponents. SER aggregates match outcomes into a global ranking that reflects relative model strength under a well-established, zero-sum rating framework. PWER extends SER by incorporating performance-related signals that reflect domination level and execution efficiency in long-horizon strategic games.



\textbf{Standard ELO Rating (SER).}
Let $R_A$ and $R_B$ denote the current ratings of model $A$ and model $B$. The expected score $E_A$ for model $A$ is defined as $E_A = \frac{1}{1 + 10^{(R_B - R_A) / \xi}}$, where $\xi$ is the logistic scale factor, conventionally set to $400$. After a match, SER is updated via 
\begin{equation}
R'_A = R_A + K \cdot (S_A - E_A),
\end{equation}
where $S_A \in {0, 0.5, 1}$ corresponds to loss, draw, or win, and $K=32$ is the update coefficient.

\textbf{Performance-Weighted ELO Rating (PWER).}
Despite its robustness and widespread adoption, SER does not differentiate between victories achieved with drastically different levels of resource consumption or strategic efficiency. Therefore, we introduce PWER, which integrates unit preservation and temporal efficiency into SER. We define a \textbf{Performance Multiplier ($M$)} that scales the reward for the victor based on their domination level:
\begin{equation}
    R'_A = R_A + K \cdot M \cdot (S_A - E_A).
\end{equation}
The multiplier $M$ is computed from two objective battle statistics:
\begin{equation}
    M = 1 + \alpha \cdot \mathcal{U} + \beta \cdot \mathcal{T},
\end{equation}
where unit preservation $\mathcal{U} = N_{\text{surviving}} / N_{\text{total}}$ measures resource efficiency, and time efficiency $\mathcal{T} = 1 - \min(T_{\text{game}} / T_{\text{max}}, 1)$ rewards decisive victories over prolonged engagements.



\subsection{Turn-Based Evaluation}
In this setting, models are allowed unlimited deliberation time, thereby isolating strategic planning and decision-making capability from inference latency. Table~\ref{tab:turn-based-results} reports the corresponding leaderboard of various LLMs.

The turn-based results exhibit a clear hierarchy that is primarily driven by reasoning capability rather than model scale. Reasoning-enhanced models consistently dominate the leaderboard, with Kimi-K2-Thinking ranking first by a large margin. The substantial PWER gap between Kimi-K2-Thinking and its instruction-tuned counterpart Kimi-K2-Instruct (371 points) provides strong evidence that explicit deliberation and long-horizon planning, rather than parameter count or instruction-following ability, are the key determinants of performance in turn-based strategic environments. A comparison between PWER and SER further reveals qualitative differences in how models achieve their outcomes. Top-performing models such as Kimi-K2-Thinking and GLM-4.7 achieve PWER scores significantly higher than their SER, indicating that their victories are not only frequent but also decisive, with high unit preservation and efficient execution. In contrast, lower-ranked models consistently suffer negative PWER adjustments relative to SER, suggesting that even when they avoid defeat, they do so at substantial cost. This ``Pyrrhic victory'' pattern highlights the limitations of win–loss–based evaluation and demonstrates the necessity of incorporating execution quality into competitive assessment.

\begin{table*}[h]
\centering
\caption{Real-Time Mode Leaderboard. Models are ranked by PWER.}
\label{tab:real-time-results}
\begin{tabular}{lccc}
\toprule
\textbf{Model} & \textbf{PWER} & \textbf{SER} & \textbf{Win Rate}\\
\midrule
GLM-4.6 & 1180.8 $\pm$ 33.7 & 1131.0 $\pm$ 16.2 & 0.75 \\
Qwen3-30B-A3B-Instruct & 1156.7 $\pm$ 44.8 & 1116.9 $\pm$ 25.4 & 0.728 \\
Amazon Nova Pro & 1103.3 $\pm$ 23.3 & 1066.6 $\pm$ 10.8 & 0.683 \\
DeepSeek-Chat & 1084.5 $\pm$ 58.1 & 1058.5 $\pm$ 38.2 & 0.621 \\
Llama-4-Scout-17B & 1073.9 $\pm$ 11.6 & 1047.0 $\pm$ 4.5 & 0.614 \\
Qwen2.5-14B-Instruct & 1064.1 $\pm$ 8.6 & 1034.0 $\pm$ 2.9 & 0.601 \\
Qwen3-14B & 1012.5 $\pm$ 27.8 & 1015.7 $\pm$ 12.9 & 0.571 \\
Claude-3.7-Sonnet & 1007.6 $\pm$ 23.5 & 1002.3 $\pm$ 11.4 & 0.540 \\
DeepSeek-R1 & 1004.2 $\pm$ 28.9 & 988.8 $\pm$ 14.4 & 0.520 \\
MiniMax-M2 & 999.9 $\pm$ 12.0 & 1005.1 $\pm$ 5.1 & 0.511 \\
DeepSeek-R1-Distill-32B & 985.0 $\pm$ 12.8 & 995.7 $\pm$ 5.7 & 0.495 \\
Qwen3-8B & 977.2 $\pm$ 2.2 & 984.8 $\pm$ 0.9 & 0.453\\
Qwen3-30B-A3B-Thinking & 972.4 $\pm$ 25.4 & 992.8 $\pm$ 11.7 & 0.418 \\
DeepSeek-R1-Distill-14B & 952.1 $\pm$ 8.5 & 971.7 $\pm$ 3.6 & 0.362 \\
Kimi-K2-Instruct & 915.3 $\pm$ 29.2 & 935.6 $\pm$ 13.4 & 0.337 \\
Qwen3-235B & 876.9 $\pm$ 12.6 & 916.0 $\pm$ 6.5 & 0.252\\
Kimi-K2-Thinking & 842.6 $\pm$ 9.5 & 892.8 $\pm$ 4.6 & 0.210 \\
Qwen3-32B & 790.9 $\pm$ 24.6 & 844.4 $\pm$ 12.2 & 0.167 \\
\bottomrule
\end{tabular}%
\end{table*}

\begin{table*}[h]
\centering
\caption{Ablation study on visual perception and logical reasoning in decision-making for VLMs and LLMs.}
\label{tab:ablation_results}
\begin{tabular}{llccc}
\toprule
\textbf{Model} & \textbf{Type} & \textbf{TCE}  & \textbf{SAE} & \textbf{Actions Per Game} \\
\midrule
GLM-4.6 & Standard LLM & 32.0\% & 29.5\% & 178 \\
GLM-4.6V & Standard VLM & \textbf{13.8\%} & \textbf{11.1}\% & 72 \\
\midrule
Qwen3-30B-A3B-Instruct & LLM(MoE) & \textbf{30.9}\%  & 33.6\% & 492 \\
Qwen3-VL-30B-A3B-Instruct & VLM(MoE) & 50.5\% & \textbf{8.2}\% & 200 \\
\midrule
Qwen3-32B & Thinking LLM & \textbf{30.2}\% & 18.9\% & 43 \\
Qwen3-VL-32B & Thinking VLM & 54.5\% & \textbf{5.3}\% & 33 \\
\midrule
Qwen3-30B-A3B-Thinking & Thinking LLM(MoE) & \textbf{18.9}\% & 32.1\% & 95  \\
Qwen3-VL-30B-A3B-Thinking & Thinking VLM(MoE) & 60.4\% & \textbf{20.0}\% & 53 \\
\bottomrule
\end{tabular}%
\end{table*}

We conducted an analysis of emergent strategic behaviors.  The qualitative analysis of model decision reveals that high-performing models exhibit distinct emergent behaviors akin to professional human players, whereas weaker models rely on simple greedy heuristics.

\paragraph{Self-Organization and Protective Rotation.} Top-tier models like DeepSeek demonstrate self-organizing defensive formations without explicit instruction. For instance, DeepSeek autonomously executed a "protective rotation": withdrawing a critically injured Archer (ID 235) to the rear while simultaneously advancing an Infantry unit (ID 232) to the front line to serve as a meat shield, explicitly describing "Infantry 232 provides frontline protection" in the Chain-of-Thought reasoning content.

\paragraph{Coordinated Strikes.} We observed coordinated attacks from the multiple agents. Kimi-K2-Thinking explicitly formulated a coordinated strike, synchronizing three agents to focus fire on a single high-value enemy unit, ignoring closer but lower-threat targets. This contradicts simple distance-based heuristics.

\paragraph{Terrain Exploitation.} Models like GLM-4.6 demonstrated environmental awareness, explicitly logging "occupying jungle terrain for defensive advantage" before engaging, leveraging the map's evasion bonuses.

These behaviors appear frequently in models with PWER $> 1100$, suggesting a phase transition where quantitative reasoning scale leads to qualitative strategic emergence.

\subsection{Real-Time Evaluation}
The results of the real-time evaluation shown in Table~\ref{tab:real-time-results} reveal a markedly different performance landscape from the turn-based setting, highlighting the critical role of inference latency in time-constrained decision-making. Models that dominated under unlimited deliberation no longer retain their advantage, and the leaderboard is reshuffled toward systems that achieve a better balance between reasoning quality and actuation speed. Notably, GLM-4.6 and Qwen3-30B-A3B-Instruct emerge at the top, while several reasoning-enhanced models, including Kimi-K2-Thinking and Qwen3-30B-A3B-Thinking, experience substantial drops in PWER. This reversal indicates that, under real-time constraints, excessive deliberation can become a liability rather than an asset.

A particularly striking contrast is observed between instruction-tuned and reasoning-enhanced variants of the same backbone. For instance, Qwen3-30B-A3B-Instruct outperforms its Thinking counterpart by nearly 200 PWER points, highlighting the performance cost of reasoning-induced latency under tight temporal constraints. Notably, this effect is not solely attributable to model architecture but is also shaped by deployment conditions: models such as GLM-4.6, when served through highly optimized inference infrastructure, retain strong reasoning capability without incurring prohibitive latency, allowing them to excel in real-time settings. Despite these differences, certain model families—most notably GLM and DeepSeek—exhibit robust cross-mode performance, suggesting that their internal reasoning strategies and inference pipelines achieve an effective balance between strategic depth and responsiveness. Overall, the real-time results reveal a trade-off that is largely obscured in turn-based evaluation: strong strategic reasoning must be coupled with timely execution to succeed when action latency becomes a first-order constraint.

The inclusion of multiple distilled models in the real-time evaluation enables a focused analysis of the effect of knowledge distillation on tactical reasoning. The results indicate that distillation largely preserves tactical decision-making capability while substantially reducing computational overhead. This efficiency gain proves particularly advantageous in time-sensitive settings, where reduced inference latency translates directly into improved real-time performance.

Another key insight from our results is the identification of a strategy–execution gap, referring to the discrepancy between a model’s ability to formulate high-quality strategies and its capacity to execute them effectively. This gap is observed across most evaluated models, indicating that strategic planning and tactical execution constitute distinct cognitive capabilities in LLMs. Notably, models with strong strategic understanding may still falter at operational execution, especially under strict time constraints. This finding underscores an important limitation of current LLM-based agents and has significant implications for deployment in time-sensitive domains, where success requires not only strategic insight but also timely and reliable action.

\subsection{Visual Perception vs. Reasoning Performance}
To disentangle the contributions of native visual grounding versus abstract logical reasoning in spatial tasks, we conduct a ablation study comparing Vision-Language Models (VLMs) against their text-only counterparts (LLMs) of identical size and architecture. We focuse on three metrics: 
\textbf{Tool Call Error (TAE)} (the frequency of the tool call errors in a game), 
\textbf{Spatial Awareness Error (SAE)} (the percentage of tool call errors explicitly caused by violations of spatial constraints, such as Manhattan distance deviation), and \textbf{Actions Per Game}.

\begin{figure}[t]
  \centering
  \includegraphics[width=0.99\linewidth]{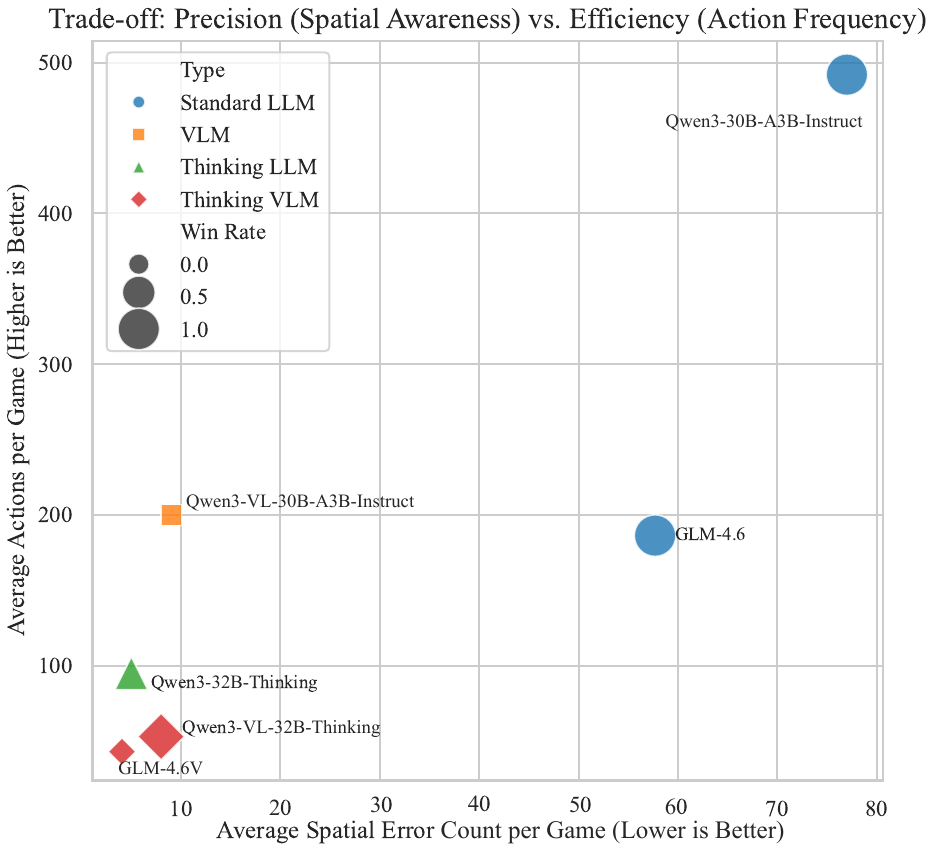}
  \caption{Trade-off analysis between Spatial Precision (x-axis) and Action Efficiency (y-axis). While VLMs (squares) achieve superior spatial grounding with minimal error, their high inference latency results in low action frequency. Conversely, standard LLMs (circles) sacrifice precision for speed. ``Thinking'' models (triangles/diamonds) bridge this gap, achieving VLM-level precision through reasoning without the visual processing overhead.}
  \label{fig:ablation_scatter}
\end{figure}

The results in Table~\ref{tab:ablation_results} and Figure~\ref{fig:ablation_scatter} reveal a fundamental perception–action trade-off in spatial decision-making. Vision-Language Models (VLMs) achieve lower spatial and tool-call errors due to explicit visual grounding, but this advantage is offset by significantly reduced action frequency caused by visual encoding latency. In the fast-paced STAR environment, execution tempo plays a decisive role: the loss of action throughput often outweighs gains in spatial precision, leading VLMs to underperform relative to faster text-only models. In contrast, standard LLMs operate at much higher action rates but exhibit pronounced spatial inaccuracies, highlighting the cost of lacking explicit perceptual grounding. 

Notably, reasoning-enhanced (``Thinking'') text-only models occupy a more favorable operating point. By leveraging explicit chain-of-thought reasoning, these models substantially reduce spatial error without incurring the latency associated with visual processing, thereby achieving a balanced trade-off between spatial accuracy and action frequency. 

\section{Conclusion}

In this work, we present the Strategic Tactical Agent Reasoning Benchmark (STAR), a comprehensive evaluation framework designed to systematically assess the reasoning depth and rapid decision-making capabilities of large language models (LLMs) in dynamic, adversarial settings. By integrating multi-agent, zero-sum game scenarios with modular, reproducible evaluation metrics, STAR provides a robust platform for quantifying strategic foresight and tactical execution across diverse models. Our extensive experiments reveal significant disparities in strategic planning, operational efficiency, and decision latency, underscoring the critical need for balanced reasoning and rapid adaptation. Notably, the observed “strategy-execution gap” suggests that models excelling in strategic reasoning may still struggle with real-time tactical execution, indicating potential avenues for model refinement. Moving forward, STAR serves as a foundational benchmark for advancing LLM evaluation in complex, evolving scenarios, promoting further research into adaptive strategic reasoning and efficient tactical decision-making.

\paragraph{Impact Statements} This paper presents work whose goal is to advance the field of machine learning. There are many potential societal consequences of our work, none of which we feel must be specifically highlighted here.

\bibliography{custom}
\bibliographystyle{icml2026}



\newpage
\appendix
\onecolumn

\section{Environment Implementation Details}
\label{sec:appendix_env}

STAR is built on a custom Entity-Component-System (ECS) engine designed for high-fidelity wargame simulation. This architecture ensures strict separation between state (Components) and logic (Systems), facilitating modularity and parallel state processing.

\subsection{Core Architecture}
The framework consists of four decoupled layers:
\begin{itemize}
    \item \textbf{Framework Layer}: Manages the ECS registry, entity lifecycle, and system scheduling.
    \item \textbf{Environment Layer}: Implements the \textit{Romance of the Three Kingdoms} ruleset, including hex-grid topology, terrain effects, and combat resolution.
    \item \textbf{Protocol Layer}: Handles JSON-based communication between the engine and agents, ensuring language-agnostic integration.
    \item \textbf{Agent Layer}: Adapts model outputs into engine-executable commands.
\end{itemize}

\subsection{Map and Terrain System}
The battlefield is a $15 \times 15$ hexagonal grid using flat-topped geometry and even-q offset coordinates. The map generator utilizes Perlin noise and cellular automata to create realistic terrain distributions.

\begin{table}[h]
\centering
\caption{Terrain Types and Effects}
\label{tab:terrain_effects}
\resizebox{0.7\columnwidth}{!}{%
\begin{tabular}{lccc}
\toprule
\textbf{Terrain} & \textbf{Move Cost} & \textbf{Defense Bonus} & \textbf{Vision Block} \\
\midrule
Plain & 1 & 0\% & No \\
Forest & 2 & +20\% & Yes \\
Hill & 2 & +30\% & Yes \\
Mountain & 3 & +50\% & Yes \\
Water & $\infty$ (Impassable) & N/A & No \\
City & 1 & +40\% & Yes \\
\bottomrule
\end{tabular}%
}
\end{table}

\subsection{Unit Mechanics and Combat Logic}
Combat resolution uses a deterministic formula modulated by unit headcount, morale, and terrain. The damage output is non-linear, modeled by a sigmoid function to simulate the "critical mass" effect in ancient warfare.

\textbf{Combat Formula:}
The effective attack power $A_{eff}$ is calculated as:
\begin{equation}
    A_{eff} = A_{base} \times \sigma(R_{count}) \times S_{status} \times T_{terrain}
\end{equation}
where:
\begin{itemize}
    \item $A_{base}$: Base attack value of the unit type.
    \item $R_{count}$: Ratio of current unit count to max count ($N_{curr}/N_{max}$).
    \item $\sigma(x)$: A sigmoid-like multiplier function defined as:
    \[ \sigma(x) = w \cdot L(x) + (1-w) \cdot P(x) \]
    where $L(x)$ is a linear decay and $P(x)$ is a polynomial decay, blending to simulate rapid combat effectiveness loss when units are decimated.
    \item $S_{status}$: Multipliers from status effects (e.g., \textit{Morale Boost}, \textit{Confusion}).
    \item $T_{terrain}$: Terrain defense modifiers.
\end{itemize}

\section{Experiment Setup}
\label{sec:appendix_setup}

\subsection{Observation Space}
Agents receive observation data in structured JSON format. The system supports multiple visibility levels, but the standard benchmark uses \texttt{FACTION} level (fog-of-war enabled).

\begin{figure}[h]
\caption{Example JSON Observation (Simplified)}
\label{lst:observation}
\begin{small}
\begin{verbatim}
{
  "faction": "shu",
  "own_units": [
    {
      "id": 101,
      "type": "cavalry",
      "position": {"col": 5, "row": 7},
      "unit_count": {"current": 95, "max": 100},
      "movement": {"current": 3, "max": 5},
      "combat": {"attack": 85, "defense": 40}
    }
  ],
  "known_enemy_units": [
    {
      "id": 204,
      "type": "archer",
      "position": {"col": 6, "row": 8},
      "estimate_count": "high" // Hidden exact numbers in FOW
    }
  ],
  "strategic_info": {
    "turn_number": 12,
    "resources": {"manpower": 850, "supplies": 400}
  }
}
\end{verbatim}
\end{small}
\end{figure}

\subsection{Action Space Reference}
The environment provides a comprehensive set of actions for agents to interact with the game world. These actions are categorized into Unit Control, Observation, Faction Control, and System operations. Below is the full specification of the supported actions, including parameters and prerequisites.

\subsubsection{Unit Control Actions}
\begin{description}
    \item[move] Displaces a unit to a specified coordinate.
    \begin{itemize}
        \item \textbf{Description}: Moves a unit to a target hex. The system performs a multi-layer validation: 1) Checks map boundaries; 2) Verifies unit ownership and turn permissions; 3) Ensures presence of required components (\texttt{HexPosition}, \texttt{MovementPoints}); 4) Calculates pathfinding cost against current Movement Points (MP), accounting for terrain penalties.
        \item \textbf{Parameters}:
            \begin{itemize}
                \item \texttt{unit\_id} (int): Unique identifier of the unit.
                \item \texttt{target\_position} (object): Target coordinates \texttt{\{col: int, row: int\}}.
            \end{itemize}
        \item \textbf{Prerequisites}: Target is reachable within current turn; Unit is not in a \texttt{CONFUSION} state; Path is not blocked by obstacles.
    \end{itemize}

    \item[attack] Engages an opposing unit in combat.
    \begin{itemize}
        \item \textbf{Description}: Initiates a combat interaction. The engine validates: 1) Faction alignment (preventing friendly fire); 2) Geometric range (Hex distance $\le$ Attack Range); 3) Resource availability (consumes 1 Action Point). It returns a detailed battle summary including damage dealt, casualties inflicted, and any terrain modifiers applied.
        \item \textbf{Parameters}:
            \begin{itemize}
                \item \texttt{unit\_id} (int): Attacker unit ID.
                \item \texttt{target\_id} (int): Target unit ID.
            \end{itemize}
        \item \textbf{Prerequisites}: Target is alive and belongs to a hostile faction; Target is within effective range.
    \end{itemize}

    \item[rest] Recovers unit condition.
    \begin{itemize}
        \item \textbf{Description}: Commands a unit to hold position, recovering Action Points and alleviating negative status effects (e.g., fatigue).
        \item \textbf{Parameters}: \texttt{unit\_id} (int).
        \item \textbf{Prerequisites}: Unit has not exhausted all actions.
    \end{itemize}

    \item[occupy] Secures a strategic tile.
    \begin{itemize}
        \item \textbf{Description}: Takes control of the current or adjacent tile, converting its ownership to the agent's faction. Essential for territory expansion and resource accumulation.
        \item \textbf{Parameters}:
            \begin{itemize}
                \item \texttt{unit\_id} (int): Unit ID.
                \item \texttt{position} (object): Target position \texttt{\{col: int, row: int\}}.
            \end{itemize}
        \item \textbf{Prerequisites}: Tile is not already owned by the faction; Unit has sufficient AP.
    \end{itemize}

    \item[fortify] Constructs defensive structures.
    \begin{itemize}
        \item \textbf{Description}: Increases the defense bonus of a friendly tile. Consumes both Action Points (AP) and Construction Points (CP).
        \item \textbf{Parameters}:
            \begin{itemize}
                \item \texttt{unit\_id} (int): Unit ID.
                \item \texttt{position} (object): Position to fortify.
            \end{itemize}
        \item \textbf{Prerequisites}: Tile is owned by faction; Fortification level is below maximum; Terrain supports construction.
    \end{itemize}

    \item[skill] Activates a special ability.
    \begin{itemize}
        \item \textbf{Description}: Triggers a unit-specific skill (e.g., "Fire Attack", "Ambush"). Validates cooldowns and resource costs (Skill Points).
        \item \textbf{Parameters}:
            \begin{itemize}
                \item \texttt{unit\_id} (int): Unit ID.
                \item \texttt{skill\_name} (string): Canonical identifier of the skill.
                \item \texttt{target} (any, optional): Context-dependent target (unit ID or position).
            \end{itemize}
        \item \textbf{Prerequisites}: Skill is unlocked and off cooldown.
    \end{itemize}
\end{description}

\subsubsection{Observation Actions}
\begin{description}
    \item[observation] Retrieves detailed observation info for a specific unit.
    \begin{itemize}
        \item \textbf{Description}: Get unit observation info.
        \item \textbf{Parameters}:
            \begin{itemize}
                \item \texttt{unit\_id} (int): Unit ID.
                \item \texttt{observation\_level} (string, optional): Observation level (default: "basic", options: ["basic", "detailed", "tactical"]).
            \end{itemize}
        \item \textbf{Prerequisites}: Unit exists.
    \end{itemize}
\end{description}

\subsubsection{Faction Control Actions}
\begin{description}
    \item[get\_faction\_state] Retrieves the overall status of the faction.
    \begin{itemize}
        \item \textbf{Description}: Get overall faction status including battles and outcomes.
        \item \textbf{Parameters}: \texttt{faction} (string): Faction name (wei/shu/wu).
        \item \textbf{Returns}: Faction state (active/victory/defeat/eliminated/draw), unit counts, and list of surviving units.
        \item \textbf{Prerequisites}: Valid faction name.
    \end{itemize}
\end{description}

\subsubsection{System Actions}
\begin{description}
    \item[end\_turn] Ends the current faction's turn.
    \begin{itemize}
        \item \textbf{Description}: Concludes the agent's turn, triggering the game engine to advance the turn counter and notify the next faction. This action restores Action Points (AP) and Movement Points (MP) for the agent's units at the start of their next turn.
        \item \textbf{Parameters}:
            \begin{itemize}
                \item \texttt{faction} (string): The faction ending its turn.
            \end{itemize}
        \item \textbf{Prerequisites}: The game is in progress and it is currently the agent's turn.
    \end{itemize}
    
    \item[get\_action\_list] Returns a schema of available actions.
    \begin{itemize}
        \item \textbf{Description}: Provides a list of supported actions and their parameter signatures, useful for dynamic agent self-correction.
    \end{itemize}

    \item[register\_agent\_info] Registers the agent with the environment.
    \begin{itemize}
        \item \textbf{Description}: Initializes the agent's session, associating it with a specific faction and recording metadata such as the model ID and provider. This must be called before performing any gameplay actions.
        \item \textbf{Parameters}:
            \begin{itemize}
                \item \texttt{faction} (string): The faction to control.
                \item \texttt{agent\_id} (string): Unique identifier for the agent.
                \item \texttt{model\_id} (string): Identifier of the LLM model.
            \end{itemize}
    \end{itemize}

    \item[strategy\_ping] Reports strategic reasoning events.
    \begin{itemize}
        \item \textbf{Description}: A telemetry action used to self-report moments of high-level strategic planning (e.g., detecting a pincer movement). These signals are aggregated to compute the Strategic Quality score.
        \item \textbf{Parameters}:
            \begin{itemize}
                \item \texttt{faction} (string): The reporting faction.
                \item \texttt{score} (float): A self-assessed confidence score (0.0 to 1.0).
                \item \texttt{evidence} (string): A text snippet describing the strategic insight.
            \end{itemize}
    \end{itemize}

    \item[report\_llm\_stats] Submits performance statistics.
    \begin{itemize}
        \item \textbf{Description}: Reports token usage, latency, and error rates to the central evaluation system at the end of a session.
    \end{itemize}
\end{description}

\subsection{System Prompt}
The system prompt guides the LLM to follow the OODA loop and utilize available tools correctly. Below is an example of the prompt for the turn-based mode.

\begin{quote}
\small
\textbf{Core Rules}

\textbf{1. Objectives \& Factions}\\
You are the strategic commander of the \textbf{\$faction\_name (\$faction)} faction. Your sole objective is to achieve victory by methodically eliminating all opposing \textbf{\$opponent\_name (\$opponent)} enemies.\\
This is a turn-based hexagonal grid wargame, so you must think and act decisively.

\textbf{2. Map \& Coordinates}\\
Map: $15 \times 15$ hex grid, using \textbf{flat-topped even-q offset} coordinates \texttt{(col,row)}.\\
Axis rules: \texttt{col} increases to the right; \texttt{row} increases upward.\\
Neighbor coordinates: \dots (hex topology rules omitted for brevity) \dots\\
Distance: Convert to axial and compute hex distance. \textbf{Forbidden}: Euclidean/Manhattan metrics.

\textbf{3. Tool Call Protocol}\\
\textbf{Exclusive Use of tool\_calls}: All actions and data requests \textbf{MUST} be executed through the \texttt{tool\_calls} field. The \texttt{content} field should only contain strategic reasoning.\\
\textbf{Mandatory Information Gathering}: \textbf{DO NOT} invent game state information. You \textbf{MUST} use tools to gather information before acting.

\textbf{4. Tools}\\
\texttt{end\_turn}: End current turn and restore AP/MP.\\
\texttt{perform\_action}: Execute an action (e.g., \texttt{move}, \texttt{attack}, \texttt{get\_faction\_state}).

\textbf{5. Recommended OODA Cycle}\\
\textbf{Observe}: Query self and enemy states.\\
\textbf{Orient}: Identify threats and opportunities.\\
\textbf{Decide}: Plan actions (e.g., attack-then-move).\\
\textbf{Act}: Call \texttt{perform\_action}.\\
\textbf{Assess}: If action fails, correct and retry.

\textbf{6. Resource Management}\\
\textbf{Action Points (AP)}: 2 AP per unit/turn. Attack costs 1 AP.\\
\textbf{Movement Points (MP)}: Move costs MP based on terrain.\\
AP and MP fully restore after \texttt{end\_turn}.

\textbf{7. Combat Mechanics}\\
Attack power scales with HP (critical mass effect). Units below 30\% HP suffer significant damage penalties. Avoid sending single units deep into enemy lines.
\end{quote}

\section{Agent System Architecture}
\label{sec:appendix_agent_design}

To evaluate Large Language Models (LLMs) across both real-time and turn-based modalities, we developed two distinct agent architectures that bridge the gap between the asynchronous game engine and the model's inference process. The system is built upon a modular Entity-Component-System (ECS) framework, governed by a central \texttt{LLMSystem} that orchestrates the interaction between the simulation state and external agents.

\subsection{Core System Components}
The \texttt{LLMSystem} acts as the primary interface for agent control, utilizing a \texttt{SyncWebSocketClient} to maintain low-latency, bidirectional communication. It delegates specific responsibilities to specialized sub-modules:

\begin{itemize}
    \item \textbf{Action Executor}: A routing layer that decouples action reception from execution. It validates incoming \texttt{ActionRequest} envelopes and dispatches them to the appropriate handler (e.g., \texttt{LLMActionHandler} for unit commands or \texttt{LLMObservationSystem} for state queries).
    \item \textbf{Telemetry \& Metrics}: The system integrates a \texttt{GameStats} component that tracks granular performance metrics, including response latency, API usage statistics, and strategic reasoning events submitted via \texttt{strategy\_ping}.
    \item \textbf{Event-Driven Notification}: The system employs a publish-subscribe model to broadcast critical game events, such as \texttt{turn\_start} and \texttt{game\_end}, ensuring agents remain synchronized with the simulation clock.
\end{itemize}

\subsection{Real-time Agent Architecture}
The real-time agent operates on a continuous-polling cycle designed to maximize responsiveness while managing inference latency. Key features include:

\begin{itemize}
    \item \textbf{Asynchronous Interaction Loop}: The agent maintains a persistent execution loop that continuously retrieves environmental observations and queries the LLM. Unlike turn-based systems, it does not await a discrete turn signal, instead acting based on the instantaneous state of the real-time environment.
    \item \textbf{Adaptive Action Throttling}: To mitigate API rate limits and simulate realistic physical constraints, the agent implements a dynamic delay mechanism. Actions incur latency proportional to their complexity; for instance, movement actions introduce delays correlated with path distance, while combat actions impose fixed animation locks.
    \item \textbf{Real-time Strategic Intent Analysis}: A background process analyzes the model's textual output stream concurrently with action execution. It employs keyword matching and sequence detection algorithms to identify strategic reasoning patterns (e.g., flanking maneuvers, encirclement tactics), reporting these to the evaluation system without impeding the primary decision loop.
    \item \textbf{Batched Command Execution}: To address HTTP request latency, the architecture supports batching multiple action calls into a single protocol envelope. This capability enables the model to execute coordinated maneuvers---such as the simultaneous movement of multiple units---within a single inference step.
\end{itemize}

\subsection{Turn-based Agent Architecture}
The turn-based agent introduces a strict synchronization layer to enforce sequential play protocols. Its design prioritizes state consistency and logical correctness over execution speed:

\begin{itemize}
    \item \textbf{Turn Synchronization Gating}: The agent employs an event-driven gating mechanism that inhibits LLM inference during the opponent's turn. The gate is lifted only upon receipt of a verified \texttt{turn\_start} event from the engine, ensuring adherence to the specific faction's turn sequence.
    \item \textbf{Temporal Context Injection}: At the onset of each new turn, the system automatically injects a state-update message (e.g., confirming turn transition and resource restoration) into the conversation history. This provides the model with explicit temporal grounding regarding state changes.
    \item \textbf{Phase Consistency Enforcement}: The architecture implements rigorous validation logic for tool invocations, ensuring the \texttt{end\_turn} command is utilized correctly and distinct from tactical actions. Execution is automatically suspended post-turn completion until the subsequent activation signal, preventing invalid actions during opposing turns.
\end{itemize}

\section{Detailed Analysis of Emergent Strategic Behaviors}
\label{sec:appendix_emergent_behaviors}

This section provides a qualitative analysis of the strategic reasoning patterns observed in top-performing models (PWER $> 1100$). By examining the decision logs from specific matches, we identified complex behaviors that emerged spontaneously from the models' general reasoning capabilities, without explicit hard-coded heuristics.

\subsection{Protective Rotation and Self-Organization}
\textbf{Model:} DeepSeek-Chat \\

In this match, DeepSeek-Chat demonstrated a sophisticated "Protective Rotation" maneuver, a tactic commonly used in professional MOBA or RTS games to preserve unit vitality.

\begin{itemize}
    \item \textbf{Context:} The Wei faction (DeepSeek) had an Archer unit (ID 235) critically injured (HP $< 20\%$) on the front line, while an Infantry unit (ID 232) was healthy but positioned in the rear.
    \item \textbf{Observed Behavior:} Instead of simply retreating the archer or attacking blindly, the model explicitly coordinated a swap. It first withdrew the archer to a safe rear position and simultaneously advanced the infantry to occupy the exact grid vacated or adjacent to the front line to block enemy vision and fire.
    \item \textbf{Reasoning Log:} The model's internal COT log recorded: \textit{"Archer 235 critically low. Executing withdrawal to (X, Y). Advancing Infantry 232 to provide frontline protection and cover retreat."}
    \item \textbf{Outcome:} The injured archer survived the match, regenerating AP/MP for later engagement, while the high-defense infantry absorbed the incoming damage.
\end{itemize}

\subsection{Coordinated Surgical Strikes}
\textbf{Model:} Kimi-K2-Thinking \\

Kimi-K2-Thinking exhibited a high-level "Focus Fire" strategy, explicitly overriding proximity-based targeting heuristics to eliminate high-threat units.

\begin{itemize}
    \item \textbf{Context:} The enemy (Shu) had a full-health Infantry unit nearby (distance 2) and a high-threat Cavalry unit further away (distance 5).
    \item \textbf{Observed Behavior:} The model calculated that the Infantry posed a low immediate threat due to lack of mobility, whereas the Cavalry could flank its archers. It ignored the nearby infantry and synchronized three separate Archer units to move into range and attack the Cavalry in the same turn.
    \item \textbf{Reasoning Log:} The chain-of-thought log showed: \textit{"Operation Lightning Strike: Ignore Infantry 236. Priority target is Cavalry 233. Unit A move to (X1,Y1), Unit B move to (X2,Y2). Focus fire to eliminate Cavalry before it acts."}
    \item \textbf{Outcome:} The Cavalry unit was eliminated in a single turn (100\% $\to$ 0\%) before it could take a turn, neutralizing the primary offensive threat.
\end{itemize}

\subsection{Environmental Awareness and Terrain Exploitation}
\textbf{Model:} GLM-4.6 \\

GLM-4.6 demonstrated a strong understanding of the ECS(Entity-Component-System) terrain modifiers, actively seeking advantageous ground.

\begin{itemize}
    \item \textbf{Context:} A skirmish in a mixed terrain area with Plains, Forest, and Hills.
    \item \textbf{Observed Behavior:} The model identified that Forest tiles provide a $+20\%$ defense bonus and vision obstruction. Before engaging in combat, it expended extra Movement Points to position its units inside Forest tiles rather than the shorter path through Plains.
    \item \textbf{Reasoning Log:} \textit{"Enemy approaching. Occupying jungle terrain at (X, Y) for defensive advantage (+20\% DEF) before engaging. This will force enemy to attack into cover."}
    \item \textbf{Outcome:} The unit took significantly reduced damage from enemy counter-attacks, winning the attrition battle.
\end{itemize}

These examples illustrate that large reasoning models are capable of:
\begin{enumerate}
    \item \textbf{State Abstraction}: Converting raw coordinates into tactical concepts like "Frontline" and "Rear".
    \item \textbf{Counter-Intuitive Optimization}: Sacrificing immediate AP/MP efficiency (e.g., taking a longer route) for long-term strategic advantage (terrain bonus).
    \item \textbf{Multi-Agent Synchronization}: Coordinating actions across distinct units to achieve a non-linear combat effect (focus fire).
\end{enumerate}




\end{document}